\documentclass{cta-author}

\usepackage{amsmath,amssymb,amsfonts}
\usepackage{algorithm}
\usepackage{algpseudocode}
\usepackage{mathtools}
\usepackage{enumerate}
\usepackage{bm}
\usepackage{booktabs}
\usepackage{threeparttable}
\usepackage{balance}

{}
{}
{}

\usepackage{soul}

\begin{document}

\supertitle{Submission for IET Radar, Sonar {\&} Navigation}

\title{A Fully-automatic Side-scan Sonar SLAM Framework}

\author{\au{Jun Zhang$^{1\correnvelope}$}, \au{Yiping Xie$^{1}$}, \au{Li Ling$^{1}$}, \au{John Folkesson$^{1}$}}

\address{\add{1}{Division of Robotics, Perception and Learning, KTH Royal Institute of Technology, Stockholm, Sweden}
\correnvelope{juzhang@kth.se}}

\begin{abstract}
Side-scan sonar (SSS) is a lightweight acoustic sensor that is frequently deployed on autonomous underwater vehicles (AUVs) to provide high-resolution seafloor images. However, using side-scan images to perform simultaneous localization and mapping (SLAM) remains a challenge when there is a lack of 3D bathymetric information and discriminant features in the side-scan images. To tackle this, we propose a feature-based SLAM framework using side-scan sonar, which is able to automatically detect and robustly match keypoints between paired side-scan images. We then use the detected correspondences as constraints to optimize the AUV pose trajectory. The proposed method is evaluated on real data collected by a Hugin AUV, using as a ground truth reference both manually-annotated keypoints and a 3D bathymetry mesh from  multibeam echosounder (MBES). Experimental results demonstrate that our approach is able to reduce drifts from the dead-reckoning system. The framework is made publicly available for the benefit of the community\footnotemark.
\end{abstract}

\maketitle

% \footnotetext{\url{https://github.com/halajun/diasss}}
\footnotetext{https://github.com/halajun/diasss}

\section{Introduction}
\label{sec:intro}

Most commercial autonomous underwater vehicles (AUVs) rely on the dead-reckoning system for underwater navigation, but this is subject to unbounded error due to accumulation of sensor uncertainties. The common solutions to limit such error are to use either global referencing system (i.e., GPS) that requires the vehicle to resurface regularly, or pre-installed acoustic ranging systems (e.g., long/short/ultrashort baseline), which are cumbersome to deploy and restrictive to limited range. An alternative solution is to incorporate sensor measurements of the environment to reduce the dead-reckoning drift using a simultaneous localization and mapping (SLAM) method~\cite{thrun2002acm}.

For AUV surveying deep undersea, the most frequently-used sensors are two types of sonars: side-scan sonar (SSS), useful for producing high resolution images of a large swath of the seafloor, and  multibeam echosounder (MBES),  useful for producing the 3D geometry of the seafloor albeit generally at a lower resolution than SSS. MBES uses beamforming to obtain directional information from the received sound waves and directly provides $3$D data of the seabed as point clouds that are relatively easy to interpret for underwater SLAM applications~\cite{palomer2016sensors}\cite{torroba2019iros}. The main drawbacks of MBES, however, are their high cost, bulky size and high power consumption. Though many compact and power-efficient MEBS models have come out in recent years, they are still very costly. SSS is more economical, lightweight and easier to deploy compared to MBES, well suited for small AUVs. It ensonifies the seafloor by emitting sound waves from the two sides (port and starboard) of the vehicle, and forms an image by stacking the pings along the vehicle's travelling direction~\cite{lurton2002springer}.  The high resolution, in principle, allows for precise AUV localization, whilst the larger coverage allows for localization using large areas of the seafloor.  

However, the problem of SLAM with SSS images is far from being solved, mainly due to the challenge of registering SSS images to one another and the lack of $3$D information. In particular, raw SSS images are geometrically distorted and unevenly ensonified, making the appearance of same region of seabed  vary in  SSS images as observed from different positions. To address this, we first apply a canonical transformation to the raw SSS images to reduce these distortions and redistribute the image pixels to be from approximately equal size patches of seabed. Then a robust salient keypoint detection and matching algorithm, combining both geometric and appearance constraints, is proposed to find keypoint correspondences between overlapping images. The inability of measuring $3$D data makes the estimate of AUV pose underdetermined, even with good associated keypoints. To mitigate such underdetermination, we initialize the depth of landmark variable with AUV's altitudes under relatively flat seafloor assumption.

With the above considerations, we propose to refine the AUV poses and landmarks of detected keypoints as a graph SLAM problem solved in two steps. First, we propose a two-ping keypoint measurement model, which is formulated together with dead-reckoning constraints as a least-squares minimization problem, to give a relative pose constraint between the two associated pings. In the second step, all of such estimated constraints are considered as loop-closure constraints in a pose graph and global optimization that refines the entire AUV pose trajectory. The ping-based solution is chosen instead of the commonly-used, computational efficient submap-based methods~\cite{aulinas2010oceans}\cite{aulinas2011icmrc} (that first group the pings into an image and then form a constraint between images based on assuming no dead-reckoning drift over the small scale of the image) due to having very sparse detected keypoints (see Fig.~\ref{fig:match_kps} left). Furthermore, we argue that our solution avoids the assumption that dead-reckoning drift can be neglected in a submap, while the computational cost  is small due to the sparseness of the constraints.

We demonstrate how our proposed method performs using a variety of metrics on actual side-scan data. This includes evaluations used by hand annotating the images but in contrast to most previous works we show automatic detection of keypoints and only use the annotated keypoints for comparison. In summary, the contributions of this work are:

\begin{itemize}
    \item We propose a fully-automatic side-scan sonar SLAM framework that is able to robustly detect keypoint correspondences and use them to refine the AUV pose trajectory from dead-reckoning system, and make it open-source for the benefit of the community;
    \item We present a feasible way of evaluating underwater SLAM using manually-annotated keypoints and a mesh built from MBES data, which are difficult to achieve with underwater data. With the proposed metrics, we are able to demonstrate precisely the performance of our proposed SLAM method against other baselines.
\end{itemize}

In the following, we show in details the methodology and implementation of our proposed framework in Sec.~\ref{sec:method}, after related work in Sec.~\ref{sec:related_work}. Experimental results are documented in Sec.~\ref{sec:experi}, with concluding remarks summarised in Sec.~\ref{sec:concl}.

\section{Related Work}
\label{sec:related_work}

Research on SLAM with SSS has drawn considerable attention within the community in the past two decades. Among the earlier works, \cite{ruiz2003oceans}\cite{reed2006tip} combine a stochastic map with Rauch-Tung-Striebel (RTS) filter to estimate AUV's location, where the stocahstic map is formulated as extended Kalman filter (EKF), with landmarks manually extracted from the side-scan images. Later in~\cite{fallon2011icra}, the authors propose to fuse acoustic ranging and side-scan sonar measurements in pose graph SLAM state estimation framework that avoids inconsistent solution caused by information lost during linearization~\cite{julier2001icra}, to achieve accurate AUV trajectory estimation. Similarly, \cite{bernicola2014oceans} proves the capabilities of using SSS images in graph-based iSAM~\cite{kaess2008tro} algorithm to produce corrected sensor trajectories for image mosaicking. Issartel et al.~\cite{issartel2017oceans} take a further step to incorporate switchable observation constraints in a pose graph to address false data association issue and guarantee robust solution. The above solutions prove the feasibility of using SSS information to help better localize AUVs, however, they rely on manual annotation of landmarks in SSS images, which goes against full autonomy in AUVs.

There are but few approaches attempting to concurrently address the problems of both SSS image association and SLAM. \cite{aulinas2010oceans} proposes to use a boosted cascade of Haar-like features to perform automatic feature detection in SSS images. The correspondences are found by means of joint compatibility branch and bound (JCBB) algorithm~\cite{neira2001tra}, and used to update pose states in an EKF SLAM framework. By applying a similar framework, \cite{siantidis2016auv} combines local thresholds and template-based detectors with strict heuristics to avoid false associations. The integration of JCBB in EKF is arguably well-fitted to solve the complete SSS SLAM problem with fewer landmark measurements, but it suffers from inconsistent performance and offers lower accuracy when comparing to graph-based SLAM solution~\cite{fallon2010ijrr}.

Meanwhile, some efforts have been made to exploit feature extraction and matching in side-scan sonar images, mainly for estimating the relative pose of AUV through image registration. \cite{vandrish2011sutwsuscrt} is one of the earliest works that try to apply feature-based registration method using SIFT~\cite{lowe2004ijcv} for side-scan sonar images. Later in~\cite{king2013oceans}, the authors extensively compares the performance of off-the-shelf feature detection techniques in matching side-scan sonar image tiles, and integrate some of the robust solutions into an image registration framework for their AUV route following system~\cite{king2012auv}. While interesting conclusions are drawn in the above works, none has integrated an applicable solution into a full SLAM framework.

Lately, the fusing of visual and geometric information to accomplish data association becomes a trending solution. For instance, \cite{mackenzie2015ccece} uses elevation gradients as additional features to assist in associations between landmarks. In~\cite{petrich2018oe}, the authors propose to construct features with $3$D location and feature intensity, and associate between them through a modified version of iterative closest point (ICP)~\cite{besl1992sf} scan matching method. In this paper, we follow the same idea and propose a robust keypoint detection and matching algorithm, which combines visual descriptors with geometric information from dead-reckoning data and sliding compatibility check to find keypoint correspondences between SSS images. The keypoints are fed into our graph-based SLAM framework to refine the dead-reckoning AUV trajectory.

Another line of work that is highly relevant is feature-based SLAM framework using forward looking sonars (FLS)~\cite{westman2018icra}\cite{li2018ral}\cite{westman2019joe}. The differences are mainly in the front-end due to different sensor modalities and the application scenarios. The typical range of FLS is from meters to over a few tens of meters, making it suitable for robust SLAM in small-scale environments, while SSS for industrial-scale surveys are generally used to map seafloor of hundreds and thousands of meters in large-scale environments. FLS usually have a wide aperture, so that there is a large amount of overlapping between swaths from consecutive frames or when there is a loop closure. Solving such an acoustic bundle adjustment is very similar to using SSS with a submap-based approach but without the need to ignore the dead-reckoning errors within submaps.

\section{Methodology}
\label{sec:method}

\begin{figure}[h]
 \centering
 \includegraphics[width=1.\columnwidth]{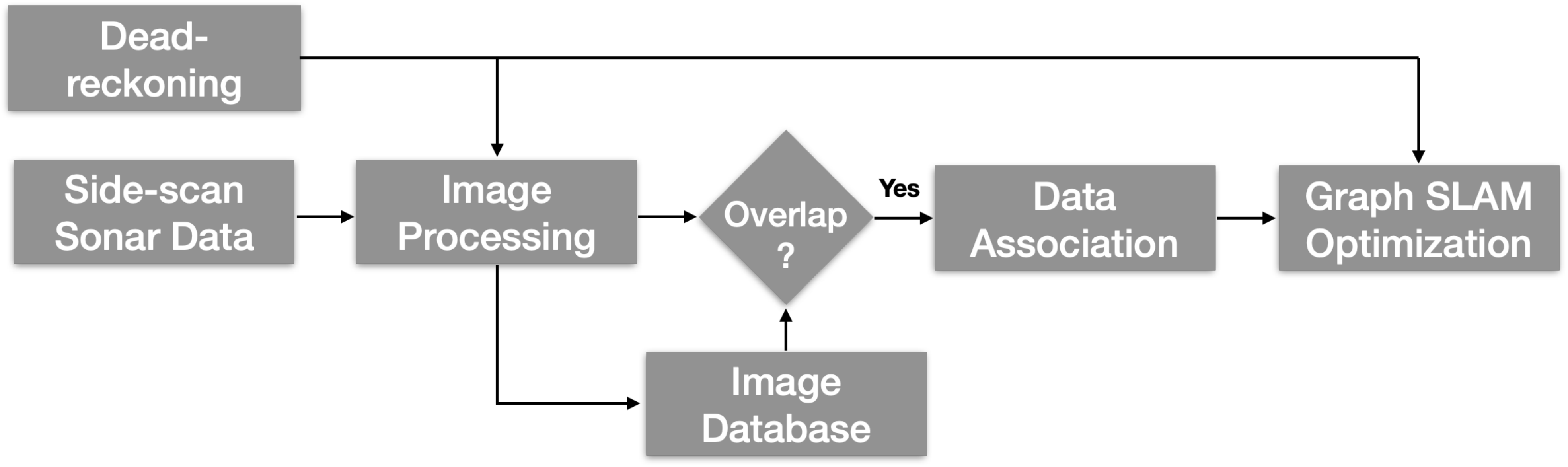}
 \caption{Overview of our SSS SLAM framework.}
 \label{fig:pipeline}
 \vspace{-.4cm}
\end{figure}

In this section, we describe the details of our proposed SSS SLAM framework, which takes pre-processed sonar image and dead-reckoning data as input, and outputs an optimized AUV pose trajectory. The proposed pipeline is summarized in Fig.~\ref{fig:pipeline}, and it contains three main parts: image processing, data association and graph SLAM optimization.

\subsection{Image Processing}

\subsubsection{SSS Image Generation}
Raw side-scan sonar data is generated continuously ping by ping along the longitudinal direction, i.e., direction of travel. To obtain an informative representation of the seabed, the time series of return intensities from individual pings are layered to form side-scan images. In our work, given that the AUV follows a common lawnmower pattern with major overlap between swaths, the proposed framework takes the side-scan image formed by each swath as input. The raw SSS pings contain over $40k$ bins per ping. This high across-track resolution, together with the low along-track resolution ($\sim$$0.5$m per ping), significantly distorts the seabed appearance. To avoid this issue, the SSS pings are downsampled to $1301$ bins per side (port and starboard).

\begin{figure}[h]
 \centering
 \includegraphics[width=.98\columnwidth]{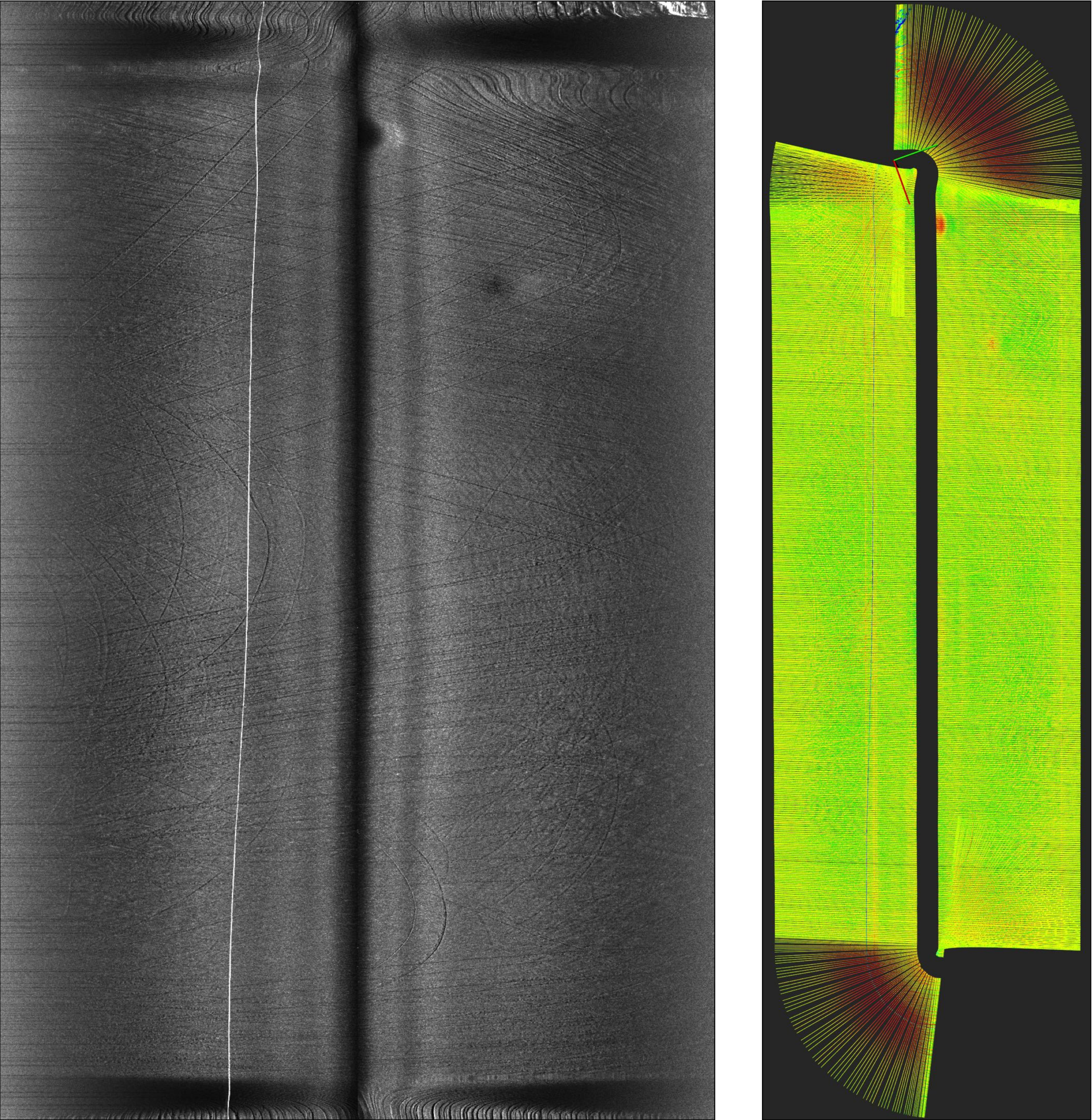}
 \caption{Left: sample of our collected SSS images after processing, where trawling marks can be observed. Right: the corresponding geo-referenced image storing pixel positions relative to the AUV poses. Note that the color encodes pixel intensity that is best viewed in the left image.}
 \label{fig:demoimg}
 % \vspace{-.4cm}
\end{figure}

\subsubsection{Canonical Transformation}
Data association across different SSS images is challenging, as the same physical area on the seabed will appear differently in the SSS images collected from different distances and angles. Furthermore, an SSS image will be distorted compared to an orthographic projection of the ensonified region and will vary in intensity as a function of the sonar position. To reduce such distortions, we apply a method~\cite{xu2023oceans} capable of transforming SSS images from different survey lines into a canonical representation, which includes two steps: intensity correction and sensor independent slant range correction. Intensity correction is performed to normalize the intensities across images based on the 'Lambertian' model~\cite{coiras2007tip}\cite{burguera2014etfa}, with the SSS backscatter intensity being proportional to the cosine square~\cite{aykin2013oceans} of the incidence angle. The slant range correction adjusts for the varying projection of side-scan pixels on the assumed horizontal seafloor. An adjustment of sine of the incidence angle is performed to be a distance along the horizontal, so that the pixels can be of fixed size in terms of horizontal range rather than slant range. More details can be found in~\cite{xu2023oceans}.

\subsubsection{Geo-reference Image}
To approximately check whether any two of the SSS images are overlapped, we utilize the dead-reckoning data to make each pixel in SSS image geo-referenced~\cite{king2012auv}, i.e., approximate the location of each pixel of an SSS image in the global reference coordinate. This is also used to narrow the searching area when performing keypoint matching in data association. An example of our collected side-scan image after image processing is demonstrated in Fig.~\ref{fig:demoimg}.

\subsection{Data Association}

Once an overlapping area is found between a new SSS image and any image in the database, a data association process is performed to find feature correspondences between the overlapping images. In this work, salient keypoints are detected as sparse features and matched between images. Specifically, evenly distributed FAST~\cite{rosten2006eccv} corners are detected in each SSS image by dividing each SSS image into grid cells and detecting corners in each cell. The reason for such an homogeneous distribution is to ensure good correspondences are found dispersedly, rather than locally in each SSS image, and thereby contribute to the optimization of most pings of the corresponding pose trajectory by adding them as constraints in the pose graph. In our previous works~\cite{xu2023oceans,weiqi2022}, the descriptor matching performance of different hand-crafted features in canonical side-scan images is compared and evaluated, including SIFT~\cite{lowe2004ijcv}, SURF~\cite{bay2006eccv}, ORB~\cite{rublee2011iccv} and BRISK~\cite{leutenegger2011iccv}. Following the conclusive results, SIFT descriptor has consistently high accuracy in descriptor matching, therefore is chosen in this paper to be extracted on the detected corners.

\begin{figure}[h]
 \centering
 \includegraphics[width=1.\columnwidth]{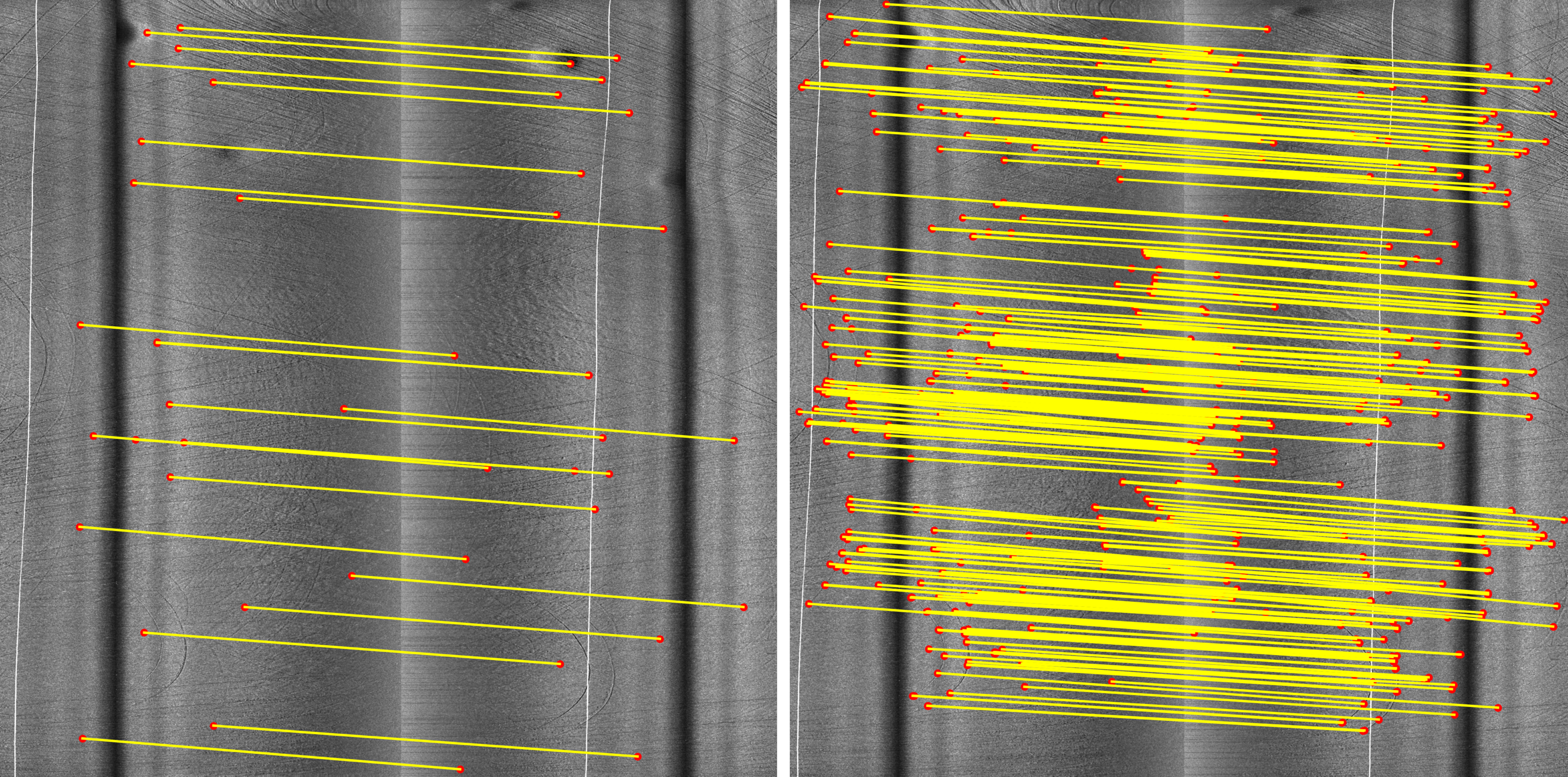}
 \caption{Example of keypoint correspondences between two adjacent side-scan images, generated by our proposed algorithm (left) and manual annotations (right). Note that the images are cropped to highlight the keypoint locations and are best viewed in color.}
 \label{fig:match_kps}
 % \vspace{-.4cm}
\end{figure}

After feature extraction is done for a new SSS image (source), the detected keypoints are used to find their correspondences in the overlapping image (target) of the database. In this paper, a simple yet effective matching approach that combines both descriptor and geometric constraints is proposed, which consists of two parts: near neighbor search and a sliding compatibility check. 

Near neighbor search uses the geo-referenced result of SSS images to find matching keypoints in local neighboring region, to avoid visual ambiguity in the case of brute-force search. For each queried keypoint in the source image, we calculate the Euclidean distance between its geo-referenced coordinate and the ones of all keypoints in the target image. Keypoints within a radius threshold $r$ are counted as matching candidates, where $r$ is set empirically depending on the geometrical distance between two images. Then the matched keypoint is found among the candidates with minimum  descriptor distance against the queried keypoint. 

The sliding compatibility check is used to verify the consistency along rows (i.e., ping) direction of the found keypoint pairs. Ideally, all keypoint pairs detected in two overlapping SSS images should have the same row difference along parallel survey lines, if we had a constant AUV speed, direction, and ping rate. In this paper, a fast RanSaC (Random Sample Consensus)~\cite{fischler1981acm} is performed by modelling the row difference as criteria, and then counting the consistency of all keypoint pairs as score to find the best inlier set. This allows eliminating many false matching pairs quickly. An example of detected keypoint correspondences is shown in Fig.~\ref{fig:match_kps} left.

\subsection{Graph SLAM Optimization}

In this section, we show how each pair of the keypoint correspondences is used to construct a measurement model and solve for the relative pose between pings of the keypoint pair. The estimated relative pose and its uncertainty serve as loop-closure constraint to correct the drift of dead-reckoning in pose graph SLAM optimization.

\subsubsection{Keypoint Measurement Model}
We model the $i^{th}$ measurement of a keypoint found in an image as a $2$D measurement that constrains the slant range to the keypoint and that it lay in a plane perpendicular to the sonar array:
\begin{eqnarray}
{\bf{z}}_{i} = \begin{pmatrix} r_i \\ 0\\ \end{pmatrix} = \hat{\bf{z}}_{i} + {\bm{\eta}}  = \begin{pmatrix} \sqrt{{\bf{\pi}}({}^{o}{\bf{x}})\cdot{\bf{\pi}}({}^{o}{\bf{x}})}  \\ (1,0,0)\cdot{\bf{\pi}}({}^{o}{\bf{x}})\\ \end{pmatrix} + {\bm{\eta}},
\label{eq:KMM}
\end{eqnarray}
\noindent where ${\bm{\eta}}$ is measurement noise, ${}^{o}{\bf{x}}\in {\rm I\!R}^3$ is the $3$D landmark on the seafloor in a global frame coordinate $o$ that is observed from this $i^{th}$ keypoint, and $r_{i}$ is the range to this landmark. ${\bf{\pi}}(\cdot)$ is a function that transforms a $3$D landmark from the global frame coordinate to the sensor frame coordinate $s$:
\begin{eqnarray}
{}^{s}\bar{\bf{x}} = {}^{b}{\bf{T}}_{s}^{-1}\cdot{}^{o}_{}{\bf{T}}_{b}^{-1}\cdot{}^{o}\bar{\bf{x}}.
\label{eq:transform}
\end{eqnarray}
\noindent Here ${}^{o}_{}{\bf{T}}_{b}\in \mathrm{SE}(3)$ is the AUV body pose\footnote{we omit the subscript '$b$' later to avoid ambiguity with ping index.} of current ping that contains the $i^{th}$ keypoint, and ${}^{b}{\bf{T}}_{s}$ is sensor offset from the AUV body to the sensor, which is normally assumed as fixed and known. Note that ${}^{o}\bar{\bf{x}}\in {\rm I\!E}^3$ denotes the homogeneous representation of ${}^{o}{\bf{x}}$.

\subsubsection{Relative Pose Estimation}
\begin{figure}[h]
 \centering
 \includegraphics[width=1.\columnwidth]{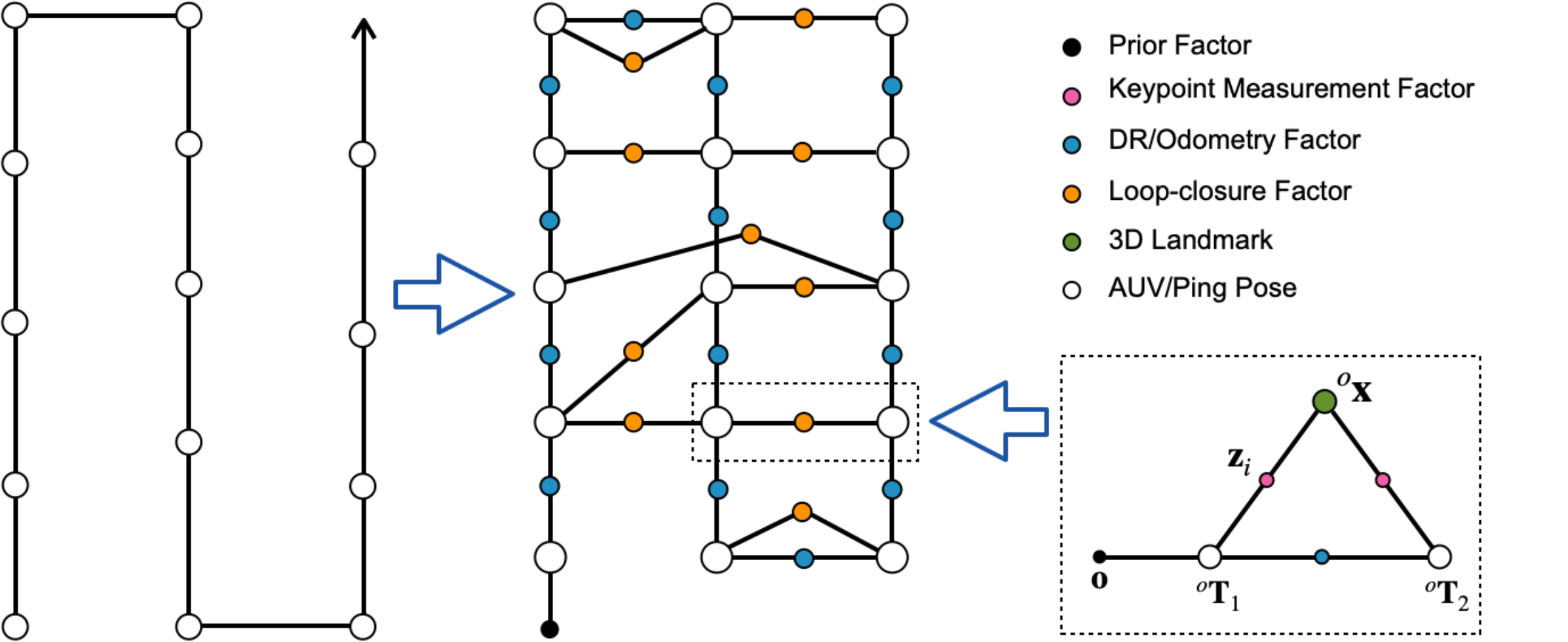}
 \caption{Factor graph representation of our proposed SSS SLAM framework. Left: an AUV surveying trajectory. Centre: factor graph of global pose trajectory. Right: factor graph for relative pose estimation.}
 \label{fig:factorgraph}
 % \vspace{-.4cm}
\end{figure}

Given a keypoint pair that observes the same $3$D landmark, the landmark and relative pose can be estimated via minimizing the minus log of least squares cost, assuming the measurement noises as Gaussian: 
\begin{eqnarray}
\begin{aligned}
& \mathcal{C}({}^{o}{\bf{T}}_{i},{}^{o}{\bf{x}})= 
\frac{1}{2} (\sum^{2}_{i=1}[(\hat{\bf{z}}_i-{\bf{z}}_i)\Sigma{}^{-1}_{i}(\hat{\bf{z} }_{i} -{\bf{z} }_{i}){}^{\top}] \\
& + [({}^{1}\hat{\bf{T}}_2-{}^{1}{\bf{T}}_2)\Sigma{}^{-1}_{t}({}^{1}\hat{\bf{T}}_{2} -{}^{1}{\bf{T}}_{2}){}^{\top}]) + \phi({}^{o}{\bf{T}}_{1}),
\end{aligned}
\label{eq:costfunc}
\end{eqnarray}
\noindent with the first term being the keypoint measurement cost discussed above, the second term the odometry measurement cost and the third term a prior cost. Specifically, ${}^{1}\hat{\bf{T}}_{2} = {}^{o}{\bf{T}}_{1}^{-1}\cdot{}^{o}{\bf{T}}_{2}$ is the relative pose between pings of the keypoints, and ${}^{1}{\bf{T}}_2$ is the measurement obtained from dead-reckoning data. The odometry term is crucial as it helps to relieve the underconstrained issue in side-scan measurement and ensure convergence to a desired minimum. $\phi(\cdot)$ is a prior model that sets one of the poses (e.g., ${}^{o}\hat{\bf{T}}_1$ here) fixed and only adjust for the other, which is reasonable as we aims to estimate the relative pose between them rather than estimate both. A factor graph describing Eq.~\ref{eq:costfunc} is illustrated in Fig.~\ref{fig:factorgraph} right.

$\Sigma{}_{i}$ is the covariance matrix for each keypoint measurement. The variance in the range measurement is dominated by the discretization of the range in the side-scan, varying from $0.01$-$1m^{2}$ depending on the resolution of the image. For the second term restricting the landmark to lie in the plane, the variance should grow as the square of the horizontal distance from the sonar. Since the sonar is mostly used at shallow grazing angles, we can instead use the range so that:
\begin{eqnarray}
\Sigma{}_{i} = \left(\begin{matrix} \sigma^{2}_{r} &0 \\ 0 &r^{2}_{i}\alpha^{2} \end{matrix} \right),
\label{eq:covariance}
\end{eqnarray}
\noindent where $\alpha$ is the beam width in radians. $\Sigma{}_{t}$ is the odometry covariance and is set proportional to the distance between the two dead-reckoning poses. Eq.~\ref{eq:costfunc} can be solved iteratively using an optimization method such as Levenberg-Marquardt algorithm, and the output relative pose is used as a loop-closing constraint for the final global optimization. 

\subsubsection{Global Optimization}

We model the SLAM problem as a factor graph as demonstrated in Fig.~\ref{fig:factorgraph} centre. The factor graph formulation is highly intuitive and allows for efficient implementation of batch~\cite{dellaert2006ijrr} and incremental~\cite{kaess2012ijrr} solvers. 

Two types of measurements are considered in the problem: the odometry measurements and the loop-closure measurements. The dead-reckoning system provides a smooth yet drifted AUV poses, which can be used to form odometry measurements between consecutive poses as a chain in the graph. The odometry measurement error ${\bf{e}}_{n}({}^{o}{\bf{T}}_{n},{}^{o}{\bf{T}}_{n+1})$ is defined as:
\begin{eqnarray}
{\bf{e}}_{n}({}^{o}{\bf{T}}_{n},{}^{o}{\bf{T}}_{n+1}) = {}^{o}{\bf{T}}_{n}^{-1}\cdot{}^{o}{\bf{T}}_{n+1} - {}^{n}{\bf{T}}^{odo}_{n+1} + {\bm{\eta}},
\label{eq:odometry}
\end{eqnarray}
\noindent where ${}^{n}{\bf{T}}^{odo}_{n+1}$ is the odometry measurement obtained from dead-reckoning data. The estimated relative pose between each pair of matched keypoints is used as a loop-closure measurement ${}^{i}{\bf{T}}^{lc}_{j}$ between the corresponding $i$ and $j$ pings, such that the loop-closure measurement error ${\bf{e}}_{m}({}^{o}{\bf{T}}_i,{}^{o}{\bf{T}}_{j})$ is denoted as:
\begin{eqnarray}
{\bf{e}}_{m}({}^{o}{\bf{T}}_i,{}^{o}{\bf{T}}_{j}) = {}^{o}{\bf{T}}_{i}^{-1}\cdot{}^{o}{\bf{T}}_{j} - {}^{i}{\bf{T}}^{lc}_{j} + {\bm{\eta}}.
\label{eq:odometry_and_lc}
\end{eqnarray}
Given that all the measurements follow Gaussian distributions, the AUV poses can be computed by minimizing the minus log of least squares cost as:
\begin{eqnarray}
\mathcal{C}({}^{o}{\bf{T}}_{n})= 
\frac{1}{2} (\sum^{N}_{n=1}[{\bf{e}}_{n}\Sigma{}^{-1}_{n}{\bf{e}}^{\top}_{n}] + \sum^{M}_{m=1}[{\bf{e}}_{m}\Sigma{}^{-1}_{m}{\bf{e}}^{\top}_{m}]),
\label{eq:globalcost}
\end{eqnarray}
where $N$ and $M$ is the number of odometry and loop-closure edges in the graph, respectively. $\Sigma{}_{n}$ is the odometry covariance that can be decided in a similar way as $\Sigma{}_{t}$, and $\Sigma{}_{m}$ is the covariance of loop-closure measurement that comes together with the estimated relative pose. Note that we omit the constant prior factor as shown in Fig.~\ref{fig:factorgraph} (centre) here to simplify the equation. We solve for Eq.~\ref{eq:globalcost} incrementally using iSAM2~\cite{kaess2012ijrr}, i.e., the whole pose graph are updated iteratively from the previous solution, when there are new measurements extracted from an input side-scan image and added to the graph. In this case we can avoid the increase of drift over a long trajectory causing the linearization to be far off that the wrong local minimal is found.

\section{Experiments}
\label{sec:experi}

\subsection{Data Collection and Annotation}

\begin{figure}[h]
 \centering
 \frame{\includegraphics[width=1.0\columnwidth]{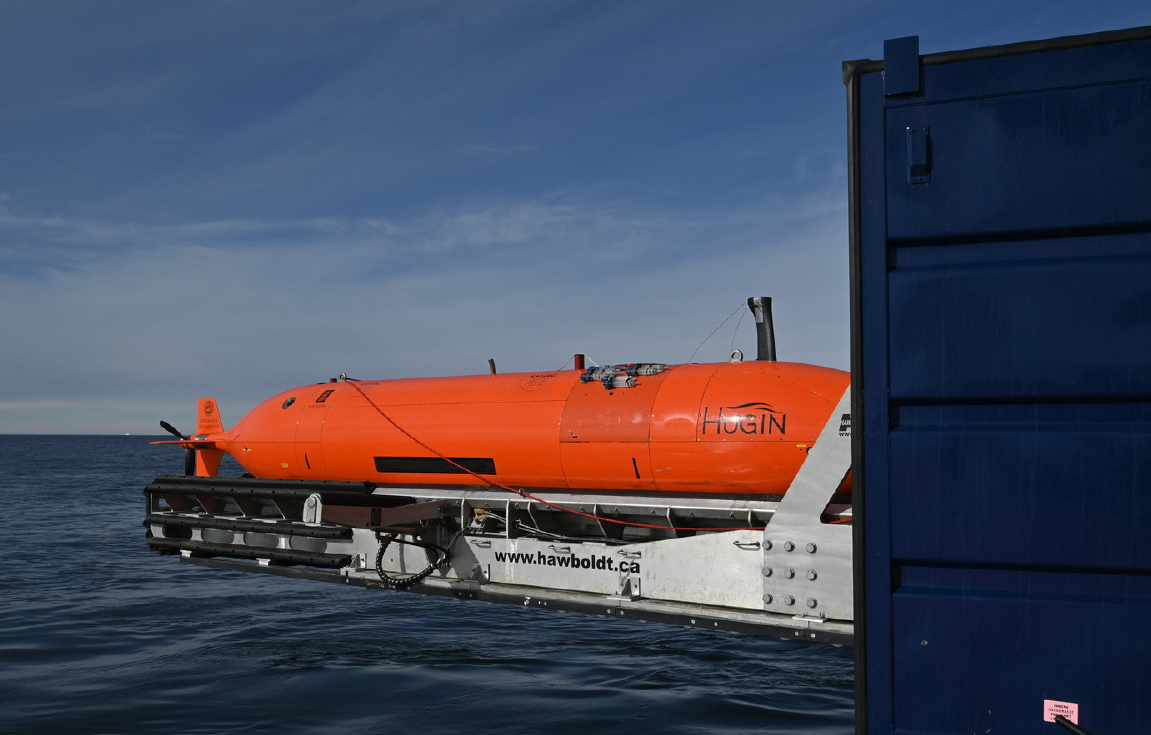}}
 \caption{The Hugin 3000 AUV used in this survey.}
 \label{fig:hugin}
 % \vspace{-.4cm}
\end{figure}

The SSS data tested in this work consist of $5$ survey lines collected in Gullmarsfjorden, western Sweden by Gothenburg University's Hugin AUV equipped with EdgeTech 2205 side-scan sonar (see Fig.~\ref{fig:hugin}). The data was collecting approximately $4$ pings per second with the AUV speed at $2$m/s.  The $5$ survey lines following a lawnmower pattern are roughly parallel to one another and have large overlap between each other. The seafloor of the surveyed area is locally flat, with lots of trawling marks. Details of the dataset and sonar characteristics can be found in Tab.~\ref{tab:dataset}.

\begin{table}[h]
  \setlength{\tabcolsep}{23pt}
  \centering
  % \fontsize{6}{7}\selectfont
  \caption{Dataset and sonar characteristics.}
  \label{tab:dataset}
 \begin{tabular}{cc}
  \toprule
                     \bf{Property}          &\bf{Value}     \cr
  \midrule
                     Bathymetry Resolution     &$0.5$m         \cr
  \midrule
                     Sidescan Type             &EdgeTech 2205         \cr
  \midrule
                     Maximum Sidescan Range            &$\sim160$m         \cr
  \midrule
                     Sidescan Frequency        &$410$KHz         \cr
  \midrule
                     Mean Altitude             &$\sim18$m         \cr
  \midrule
                     Survey Area               &$\sim200$x$800$m         \cr
  \midrule
                     Side-scan Pings           &$\sim9600$         \cr
  \bottomrule
  % \vspace{-.4cm}
 \end{tabular}
\end{table}

We manually annotate sets of keypoints between each pair of the overlapped images for evaluation purpose. The annotation process is conducted efficiently with the help of a $3$D mesh of the bathymetry that is constructed from MBES data. The $3$D mesh is used to find an initial guess of the potential corresponding keypoints when the annotator identifies a keypoint in one SSS image. Thus, given the proposed correspondences, the annotator only needs to inspect and confirm the correctness of the correspondence based on the image appearance. Using this method, we obtain about $250$$\sim$$500$ keypoint correspondences for each overlapping image pairs, see Fig.~\ref{fig:match_kps} right and Tab.~\ref{tab:keypoint}.

\begin{table}[h]
  \centering
  \fontsize{7}{8}\selectfont
  \caption{Amount of annotated and detected keypoint correspondences.}
  \label{tab:keypoint}
 \begin{tabular}{lcccccccccc}
  \toprule
  Image Pair   &0-1   &0-2   &0-3   &0-4   &1-2   &1-3  &1-4   &2-3   &2-4   &3-4   \cr
  \midrule
  Annotated       &413   &289   &274   &275   &462   &337  &306   &391   &340   &420   \cr
  Detected        &43    &25    &54    &23    &49    &22   &24    &53    &45    &38   \cr
  \bottomrule
 \end{tabular}
 % \vspace{-.4cm}
\end{table}

\subsection{Dead-reckoning Baseline}

We use navigation data from the inbuilt dead-reckoning system of Hugin as baseline reference for comparison. The dead-reckoning solution embedded in the Hugin AUV is a high accuracy Doppler Velocity Log (DVL) aided Inertial Navigation System (INS) that can integrate various forms of positioning measurements from Inertial Measurement Unit (IMU) in $1$ nmi/h class, DVL, compass and pressure aiding sensor, etc., in an error-state Kalman filter and smoothing algorithm to estimate position, velocity and attitude. The overall accuracy would be around $0.08\%$\footnote{https://www.gu.se/en/skagerak/auv-autonomous-underwater-vehicle} of the distance travelled. More specific details can be found in~\cite{jalving2003oceans,gade2003navlab}.

\subsection{Results}

\begin{figure}[h]
 \centering
 \includegraphics[width=1.0\columnwidth]{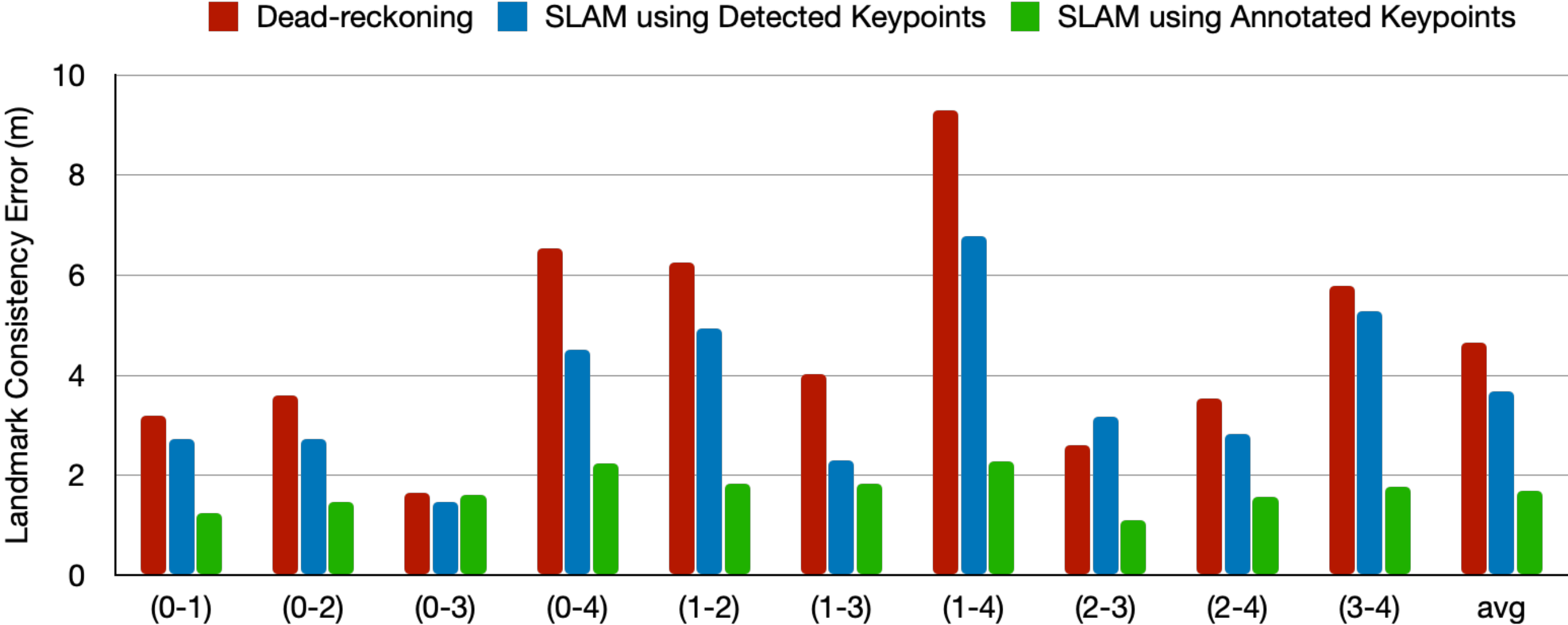}
 \caption{Landmark consistency errors that are averaged on all the annotated keypoint correspondences between each SSS image pair and over all pairs. Note that the X axis represents the index of each image pair.}
 \label{fig:lm_errors}
 % \vspace{-.4cm}
\end{figure}

In principle, a landmark observed from different ping pose should have a unique global position in ideal scenario. Based on this criteria, we propose to compute the \textit{landmark consistency error} using annotated keypoints and the bathymetry mesh from MBES data. In particular, for each keypoint pair, we project both keypoints onto the mesh by ray-casting and find their intersections with the mesh~\cite{bore2022joe}, respectively. Then the Euclidean distance between both intersected landmarks is used as the error metric, i.e., the smaller this consistency error it is, the more accurate the ping poses they are. Results are shown in Fig.~\ref{fig:lm_errors}. We can see that our SLAM estimation using annotated keypoints shows consistent error reductions in all image pairs, with an average error dropping by $2.9$m ($63\%$). Similar trend of improvements can also be seen in our SLAM estimation using detected keypoints ($21\%$ overall), except for poses between image pair $2$-$3$. 

\begin{figure}[h]
 \centering
 \includegraphics[width=1.0\columnwidth]{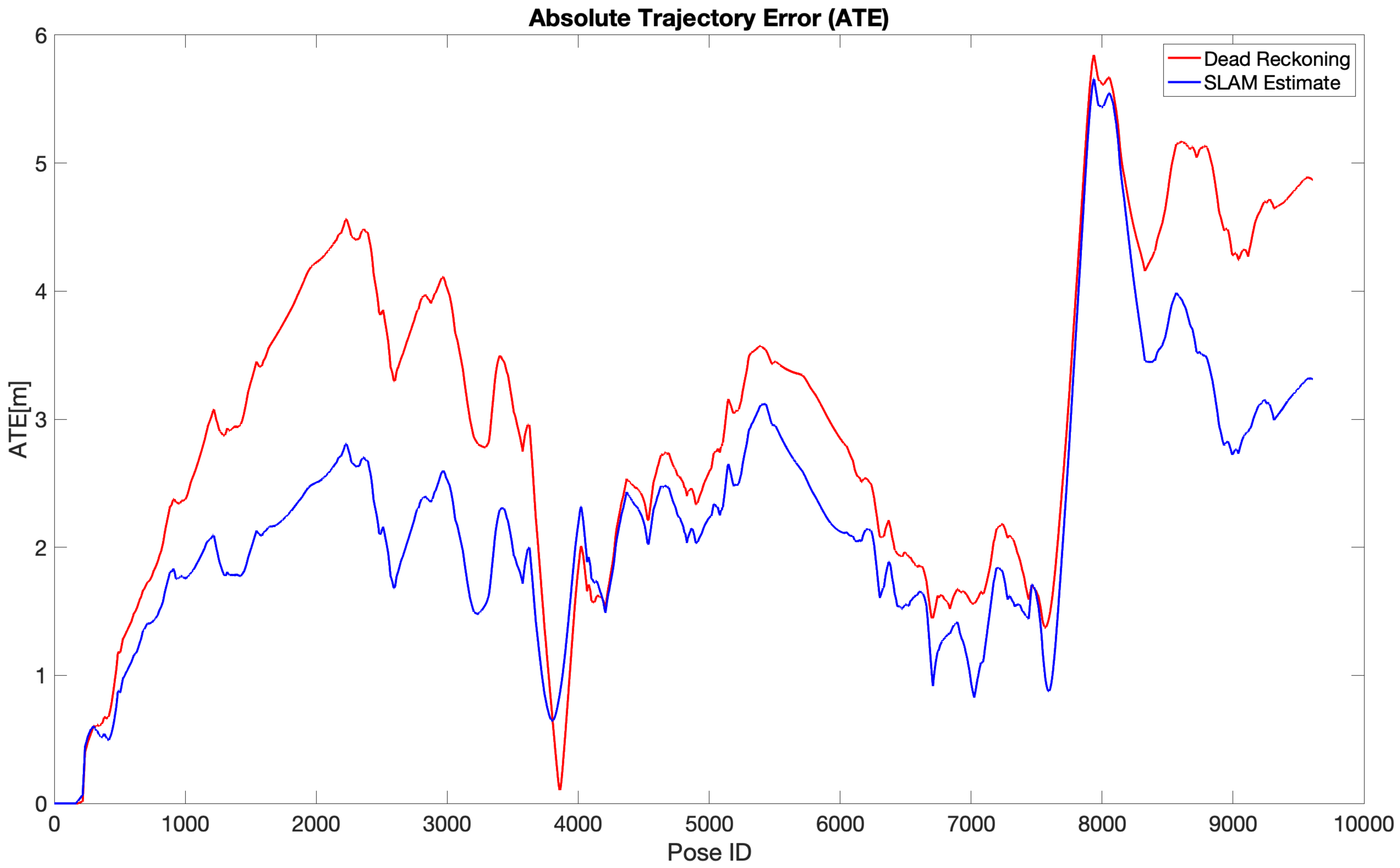}
 \caption{Comparison of absolute trajectory error (ATE) between SLAM estimation using annotated keypoints and dead-reckoning trajectory.}
 \label{fig:ate}
 % \vspace{-.4cm}
\end{figure}

\begin{figure*}[h]
 \centering
 \includegraphics[width=1.9\columnwidth]{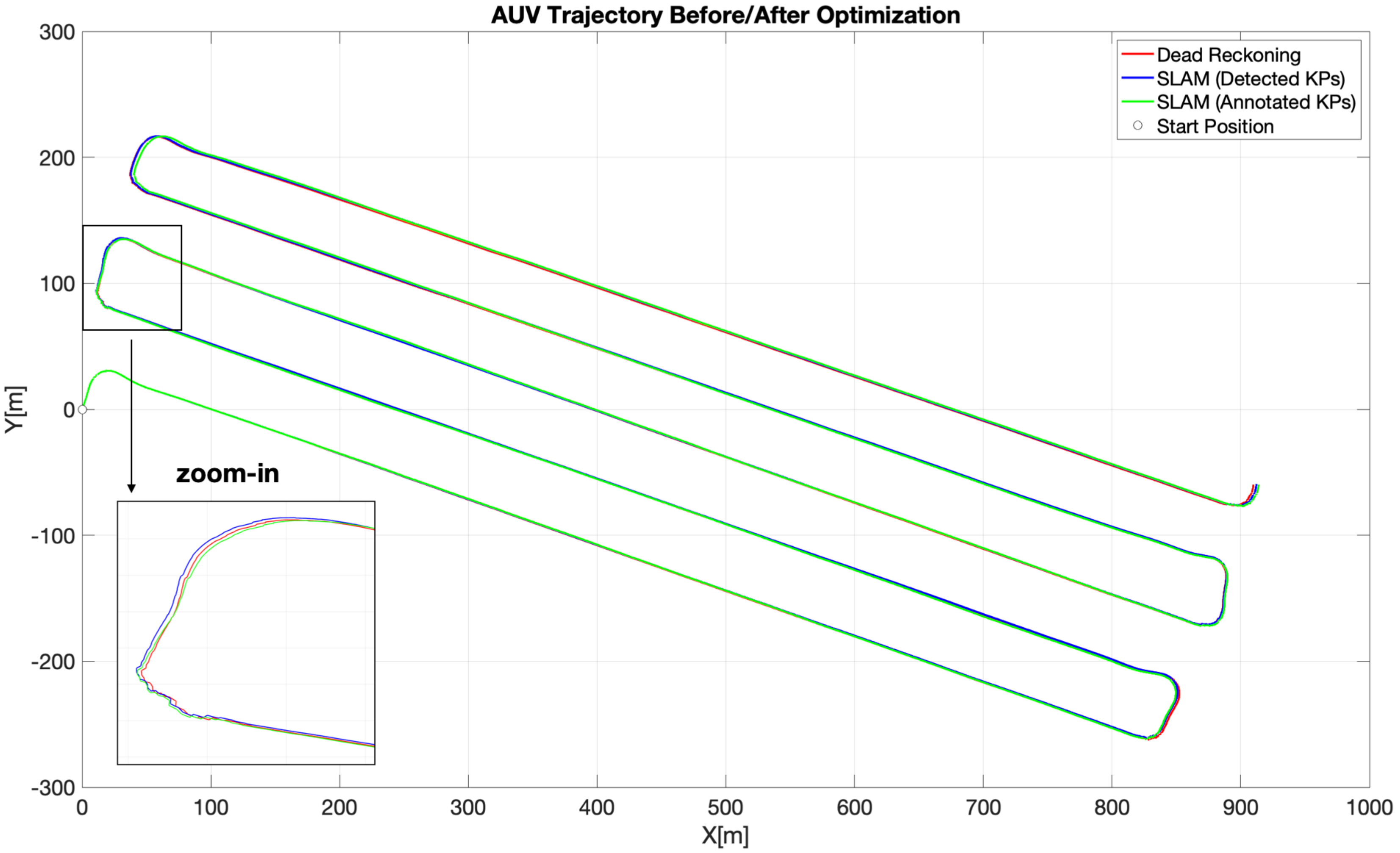}
 \caption{Comparison of absolute trajectory error (ATE) between SLAM estimations and dead-reckoning trajectory.}
 \label{fig:auv_trj}
 % \vspace{-.4cm}
\end{figure*}

We also compute the absolute trajectory error (ATE)~\cite{sturm2012iros} to evaluate absolute pose consistency, by treating the estimated pose trajectory using annotated keypoints as 'ground truth' baseline. Fig.~\ref{fig:ate} demonstrates the error comparison of the whole trajectory, where our estimated trajectory has less errors at most pings, except for the ones between $3.9$-$4.3$k, where the AUV turns back, and there is no detected keypoint constraint for corrections, as can be seen in Fig.~\ref{fig:auv_trj} the zoom-in window. Overall, the estimated trajectory is closer to the 'ground truth' trajectory, with an $25\%$ improvement over dead-reckoning in root mean squared error (RMSE), see Tab.~\ref{tab:rmse}. Interestingly, we can notice that the trajectory error with dead-reckoning does not always grow linearly over time, as Hugin running a lawnmower pattern could cancel out the drift growth obtained from body-fixed velocity and heading errors~\cite{jalving2003toolbox}. Similar effect can be observed in the landmark consistency error with dead-reckoning in Fig.~\ref{fig:lm_errors}.

\begin{table}[h]
  \centering
  \setlength{\tabcolsep}{13pt}
  % \fontsize{6}{7}\selectfont
  \caption{Comparison of ATE between Dead-reckoning and SLAM estimation.}
  \label{tab:rmse}
 \begin{tabular}{ccc}
  \toprule
                     &Dead-reckoning          &SLAM Estimation      \cr
  \midrule
  $E_{ate}$ (m)        &$3.2668$     &\bm{$2.4414$}         \cr
  \bottomrule
 \end{tabular}
 % \vspace{-.2cm}
\end{table}

\begin{figure}[h]
 \centering
 \includegraphics[width=1.0\columnwidth]{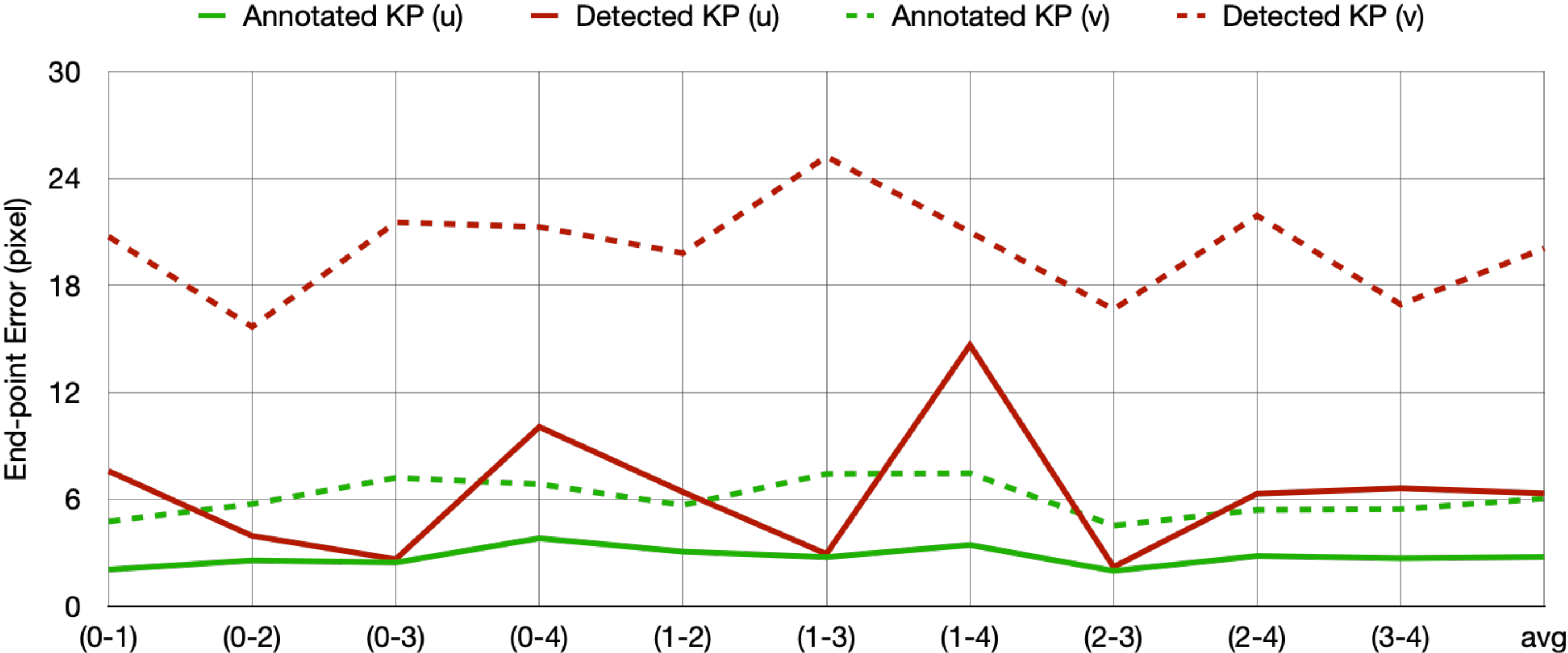}
 \caption{Averaged end-point errors of annotated and detected keypoint correspondences between each SSS image pair and over all pairs. Here $u$ and $v$ denote longitudinal and lateral direction of the image, respectively.}
 \label{fig:kp_errors}
 % \vspace{-.4cm}
\end{figure}

To evaluate the quality of detected keypoint correspondences, we make use of the $3$D mesh in a similar way as the landmark consistency metric. Specifically, for each pair of evaluated keypoint correspondence, we project one (e.g., from source image) onto the mesh and find the intersected landmark. Then we search in the target image for a keypoint that has the same intersected landmark on the mesh or close enough ($<$$0.3$m), and treat it as a baseline correspondence. We compute the end-point error (EPE)~\cite{sun2014ijcv}, i.e., the pixel distance between the detected and baseline correspondence as metric. Note that the baseline we calculate does not represent the true value, since it is searched by ray-casting using ping poses and $3$D mesh that are not absolute accurate, but we could observe the relative comparison, especially when the EPE of annotated keypoints are also added to compare, as shown in Fig.~\ref{fig:kp_errors}. We can see that our detected keypoints performs well in $u$ (longitudinal) direction, where errors at some image pairs are pretty close to those of annotated keypoints. This indicates the sliding compatibility check in pings does help increase the matching accuracy along $u$ direction. However, the accuracy in $v$ (lateral) direction is much worse due to having no constraint to regularize.

\subsection{Discussion}

\begin{figure}[h]
 \centering
 \includegraphics[width=.98\columnwidth]{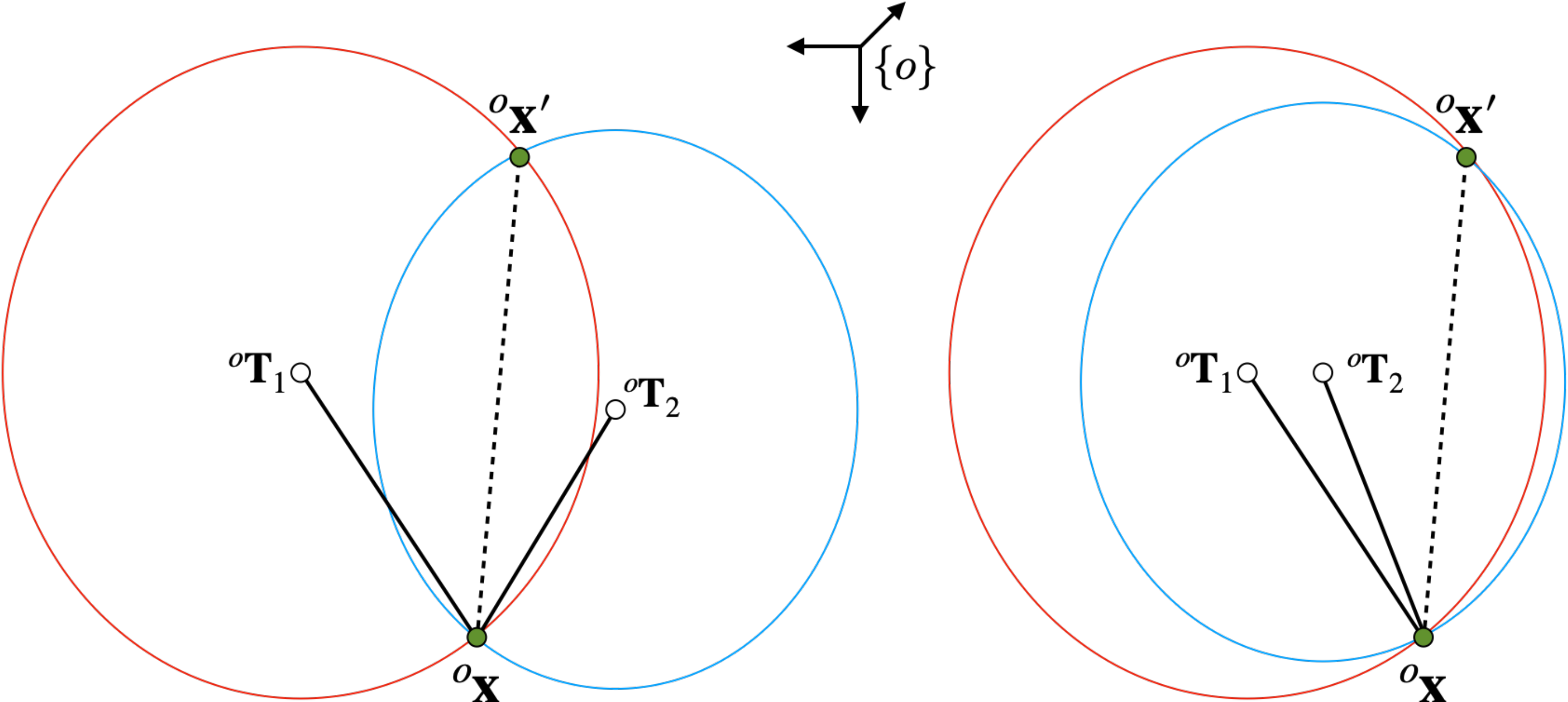}
 \caption{Illustration of degeneracy due to landmark depth ambiguity. Left: two pings observing a landmark in between them. Right: two pings observing a landmark on the same side.}\textbf{}
 \label{fig:degenerate}
 % \vspace{-.4cm}
\end{figure}

When solving Eq.~\ref{eq:costfunc} for relative pose and landmark, we discover a degenerate case of landmark depth ambiguity as shown in Fig.~\ref{fig:degenerate}, where the two circles are intersected in a line segment. The solution of landmark could be either end of this line segment (i.e., ${}^{o}{\bf{x}}$ or ${}^{o}{\bf{x}^{\prime }}$) under range constraints, and the converging direction is highly relevant to the initial values of the estimated landmark. We believe this ambiguity, together with measurements uncertainties, makes the solver prone to non-convergence or convergence towards an undesired minimum. Adding a prior on the landmark depth could help relieve from this issue. We thereby use geo-referenced images to initialize the $x$ and $y$ values of the estimated landmark. For $z$ (depth), we use the depth of the nearer ping pose as a prior, and set its uncertainty that is proportional to the Euclidean distance between its approximated position and position of this ping pose. 

\begin{figure}[h]
 \centering
 \includegraphics[width=1.0\columnwidth]{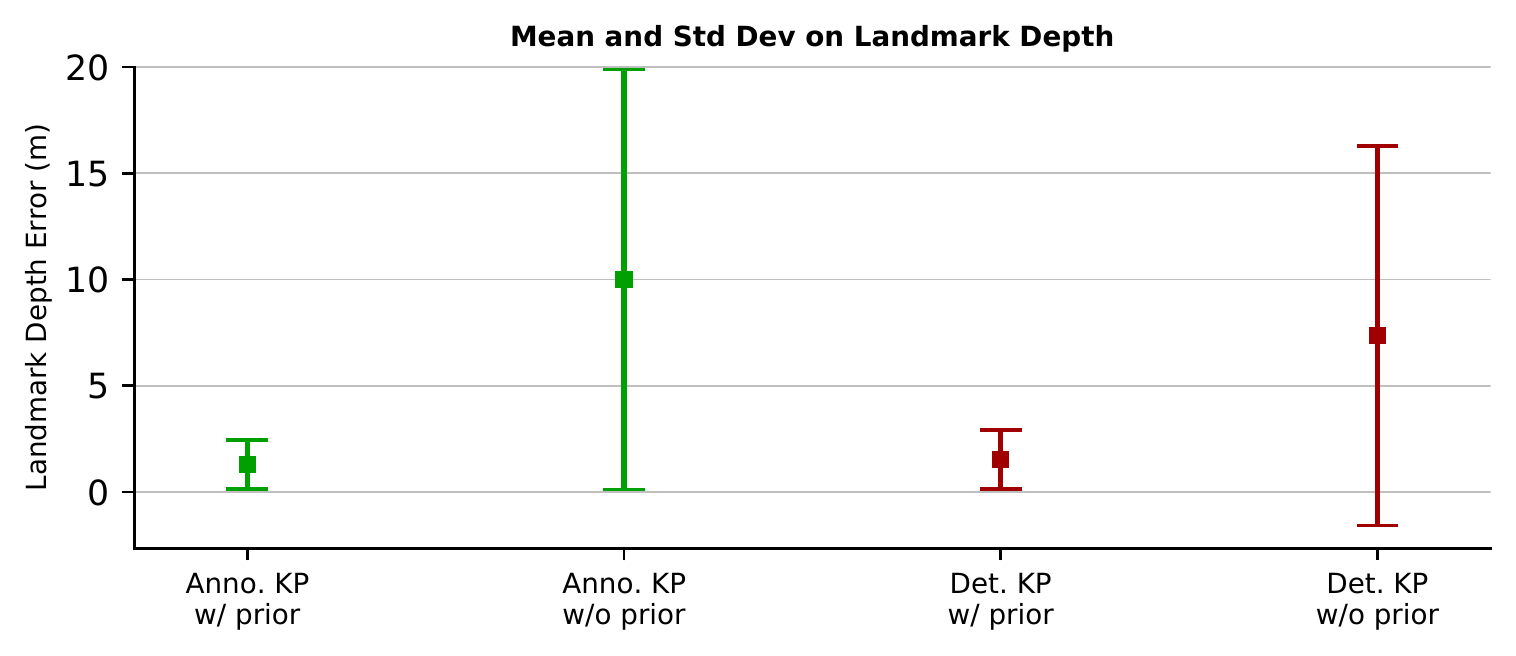}
 \caption{Mean and standard deviation of landmark depth error with (w/) and without (w/o) depth prior over all the estimated landmarks using annotated and detected keypoints, respectively. The depth error is computed using the $3$D bathymetry mesh from MBES as reference.}
 \label{fig:depth_error}
 \vspace{-.4cm}
\end{figure}

Fig.~\ref{fig:depth_error} shows the comparison of landmark depth estimation with and without depth prior. We can see that the depth errors become variably large without depth prior, while adding depth prior makes the estimate more consistent and accurate. Note that this prior only works fairly well for a relatively flat seafloor with slight sloping like the tested data in this paper, but not for uneven seabed with rich hills, holes and rocks. In that case, a high precision $3$D mesh from MBES or interferometric side-scan sonar~\cite{kolouch1984ihr} could provide strong prior to reduce this depth ambiguity, but this is out of scope of this paper.

\section{Conclusion}
\label{sec:concl}

In this paper we present a feature-based SLAM framework using side-scan sonar to improve the AUV pose trajectory from dead-reckoning data. The proposed framework automatically detects keypoint correspondences between SSS images and formulates them as constraints to refine the pose trajectory through graph optimization. We carefully test and analyse our method on real data collected by Hugin AUV, and demonstrate its effectiveness. Experimental results show that the proposed method is able to effectively reduce the drift in dead-reckoning trajectory with only side-scan sensor measurements. 

Though a robust keypoint matching algorithm is proposed in this paper, there is much space to improve quantitatively and qualitatively, as shown in Tab.~\ref{tab:keypoint} and Fig.~\ref{fig:kp_errors}. The main issue lies in not being able to detect rich repetitive keypoints that can be tracked across side-scan images. This could be possibly solved by data-driven methods~\cite{sarlin2020cvpr}\cite{sun2021cvpr}, which would be an interesting future direction, while to achieve that, more annotated keypoints are needed as training data.

\section*{ACKNOWLEDGMENT}

\noindent This work is supported by Stiftelsen fr Strategisk Forskning (SSF) through the Swedish Maritime Robotics Centre (SMaRC) (IRC15-0046) and the Wallenberg AI, Autonomous Systems and Software Program (WASP) funded by the Knut and Alice Wallenberg Foundation.

\balance
\bibliographystyle{iet}
\bibliography{refs/main}

\end{document}